\pdfoutput=1

\documentclass[11pt]{article}

\usepackage[final]{acl}

\usepackage{times}
\usepackage{latexsym}
\usepackage{amsmath}
\usepackage{amssymb}
\usepackage{CJKutf8}
\usepackage{arydshln}
\usepackage{ulem}
\usepackage{soul}
\usepackage{subcaption}
\usepackage{cleveref}
\usepackage{soul}
\usepackage{xcolor}
\usepackage{transparent}

\newcommand{\ctext}[3][RGB]{%
  \begingroup
  \definecolor{hlcolor}{#1}{#2}\sethlcolor{hlcolor}%
  \hl{#3}%
  \endgroup
}

\usepackage[T1]{fontenc}

\usepackage[utf8]{inputenc}

\usepackage{microtype}

\usepackage{inconsolata}

\usepackage{graphicx}

\title{Repetition Neurons: How Do Language Models Produce Repetitions?}

\author{
    Tatsuya Hiraoka \quad\quad Kentaro Inui \\
    Mohamed bin Zayed University of Artificial Intelligence (MBZUAI)   \\
    RIKEN\\
    \texttt{\{tatsuya.hiraoka, kentaro.inui\}@mbzuai.ac.ae}
}

\begin{document}
\maketitle
\begin{abstract}
This paper introduces \textbf{repetition neurons}, regarded as ``skill neurons'' responsible for the repetition problem in text generation tasks.
These neurons are progressively activated more strongly as repetition continues, indicating that they perceive repetition as a task to copy the previous context repeatedly, similar to in-context learning.
We identify these repetition neurons by comparing activation values before and after the onset of repetition in texts generated by recent pre-trained language models.
We analyze the repetition neurons in three English and one Japanese pre-trained language models and observe similar patterns across them.
\end{abstract}

\section{Introduction}
While text generation with LLMs such as GPT-3~\cite{brown2020language} has been actively studied, the issue of repetition remains a fundamental challenge~\cite{li2023repetition,ivgi2024loops}. 
Specifically, repetition is particularly problematic under greedy generation, which is often used when reproducibility must be guaranteed~\cite{song2024good}.

Many researchers have tackled this problem by analyzing repetition~\cite{fu2021theoretical,xu2022learning} and developing techniques to mitigate repetitive outputs~\cite{keskar2019ctrl,shirai2021neural,zhu2023penalty,li2023repetition}. 
Some works specifically focus on attention heads, such as induction heads, framing repetition as a key mechanism for in-context learning~\cite{olsson2022context,bansal2023rethinking,crosbie2024induction}. 
However, the internal mechanisms of generative models that produce repetitive outputs remain insufficiently explored~\cite{vaidya2023humans,wang2024mitigating}.

We focus on the neurons of Transformer language models~\cite{vaswani2017attention,geva2021transformer,dai2022knowledge,chen2024journey} that detect repetition in inputs and trigger repetitive outputs in text generation.
We refer to these neurons as ``\textbf{repetition neurons}'' following \newcite{wang2024mitigating}.
We hypothesize that the repetition neuron is a ``skill neuron''~\cite{radford2017learning,wang2022finding} that prompts the model to generate repetition as a task of copying the previous context, akin to ``task vectors''~\cite{hendel2023context} found in in-context learning~\cite{brown2020language,yan2024understanding}.

\begin{figure}[t]
\centering
\includegraphics[width=7.5cm]{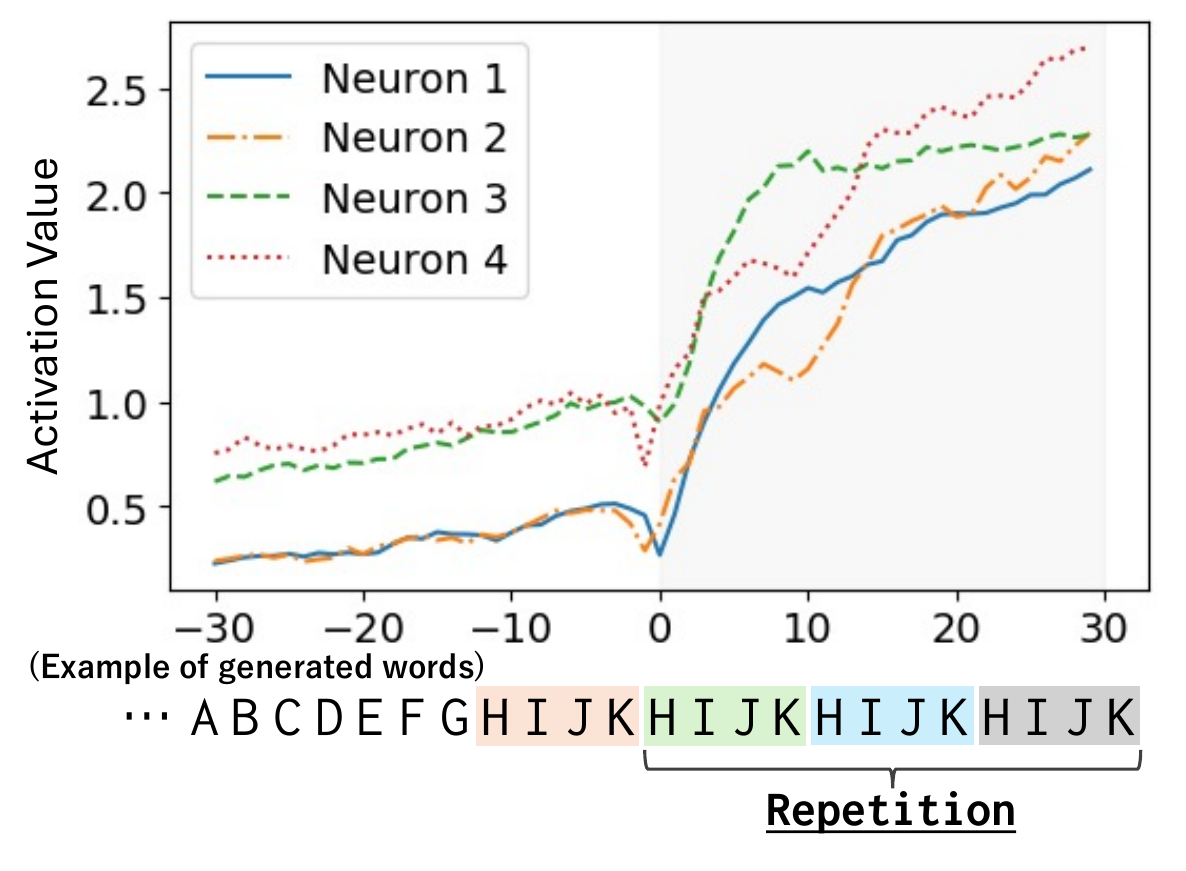}
\vspace*{-0.3cm}
\caption{
    Activation values of top four repetition neurons for 30 tokens before and after repetition (Gemma-2B, averaged value over 1,000 texts).
    \textbf{Repetition neurons are strongly activated in the repetition range.}
}
\label{fig:averagedActivation}
\end{figure}

We propose a method to identify repetition neurons by comparing activation values in the input ranges before and after the onset of repetition (\S \ref{sec:method}).
As shown in Figure \ref{fig:averagedActivation}, repetition neurons tend to become progressively more strongly activated as the repetition sequence continues.

We inspected repetition neurons in three English and one Japanese pre-trained language model. 
Our experimental results show that repetition neurons appear in both intermediate and final layers (\S \ref{sec:observation}). 
Furthermore, we demonstrate that deactivating these neurons suppresses the output probabilities of repeated tokens (\S \ref{sec:prevention}), while activating them increases these probabilities (\S \ref{sec:invoking})\footnote{Code for our experiments is available at \url{https://github.com/tatHi/repetition_neuron}}.
In addition, we highlight the relationship between repetition neurons and induction heads (\S \ref{sec:heads}).

\section{General Setting}
\subsection{Models}
\label{sec:models}
We utilized three English pre-trained language models: Gemma-2B~\cite{team2024gemma}, which has 18 layers (294,912 neurons), Pythia-2.8B-Deduped~\cite{biderman2023pythia}, with 32 layers (327,680 neurons), and LLaMA-3.2-3B~\cite{dubey2024llama}, with 28 layers (229,376 neurons).
Additionally, we employed a Japanese pre-trained language model: LLM-jp-3-1.8B~\cite{llmjp2024llm}, which has 24 layers (172,032 neurons).

\subsection{Dataset with Repetition}
\label{sec:dataset}
To analyze the internal workings of language models on repetitive text, we collected 1,000 texts containing repetition from each language model.
We randomly generated the first ten tokens with \verb|temperature = 1.0| using the \verb|generate()| method from HuggingFace Transformers~\cite{wolf2020transformers}.
Afterward, we filtered out texts that did not contain repetition.
We defined a text as containing repetition if the same 10-gram token sequence appeared three times at equal intervals within 100 tokens.
Additionally, we excluded texts that did not have at least 50 tokens before and after the onset of repetition.
The onset of repetition is defined as the point where the repeated sequence appears for the second time (see Figure \ref{fig:averagedActivation}).
Table \ref{tab:examples} in Appendix \ref{sec:example_of_dataset} provides examples of the repetitive texts generated through this process.
The entire generation process took less than two hours on a single NVIDIA V100 GPU.

\begin{figure}[t]
\centering
\includegraphics[width=7.5cm]{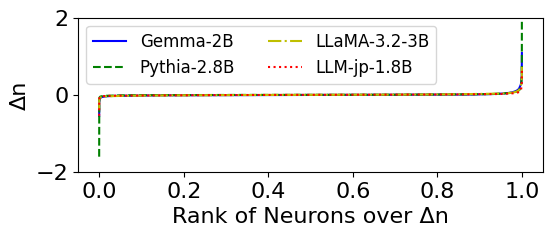}
\caption{
    $\Delta_n$ of all neurons sorted in the ascending order.
    The x-axis shows the relative rank of each neuron (i.e., 1.0 is the 294,912-th neuron in Gemma-2B).
}
\label{fig:delta}
\end{figure}

\begin{figure}[t]
\centering
\includegraphics[width=7.5cm]{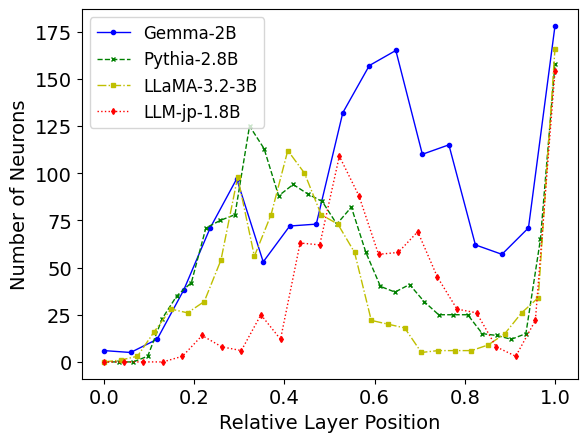}
\caption{
    The number of repetition neurons for each layer when considering the top 0.5\% of the entire neurons are repetition neurons. The x-axis shows the relative location of layers against the number of entire layers (e.g., 1.0 is the 18th layer in the case of Gemma-2B). 
}
\label{fig:neuronInLayers}
\end{figure}

\section{Finding Neurons Invoking Repetition}
\subsection{Detecting Repetition Neuron}
\label{sec:method}
In this work, we consider the outputs of the activation function in the feed-forward network of each Transformer layer as ``neurons,'' following previous studies~\cite{geva2021transformer,dai2022knowledge,wang2022finding}.
We hypothesize that repetition neurons are more strongly activated in the range of texts with repetition and less active in texts without repetition.
Therefore, we identify repetition neurons by comparing their activation values before and after the onset of repetition.

Let $x \in X$ represent a single text containing repetition, generated as described in \S \ref{sec:dataset}, with $|X|=1,000$ texts in total.
Each text consists of a sequence of $M$ tokens, $x = \{w_1, ..., w_m, ..., w_M\}$, and includes a repetition onset point $s$.
This means the sequence after position $s$ (i.e., $x_{s \leq m} = \{w_{m=s}, ..., w_M\}$) consists of repeated tokens.
We define the $r$ tokens preceding the onset point, $x_{s-r}^{s-1}= \{w_{s-r}, ..., w_{s-1}\}$, as the ``normal'' range without repetition, and the $r$ tokens following the onset point, $x_{s}^{s+r-1}= \{w_{s}, ..., w_{s+r-1}\}$, as the ``repetition'' range.
We used the hyperparameter $r=30$ for the main experiments, and Appendix \ref{sec:ablation} reports the ablation study.
For each neuron $n$ involved in the forward computation of the language model, we compute the average activation values $a_n$ and $\bar{a}_n$ over both the normal and repetition ranges, respectively.
\begin{align}
a_n = \frac{1}{|X| \times r}\sum_{x \in X}{\sum_{m=s-r}^{s-1}{f(w_m, x_{1}^{m}, n)}}, \\
\bar{a}_n = \frac{1}{|X| \times r}\sum_{x \in X}{\sum_{m=s}^{s+r-1}{f(w_m, x_{1}^{m}, n)}},
\end{align}
where $f(w_m, x_{1}^{m}, n)$ is a function that returns the activation value of neuron $n$ at the time step corresponding to the input token $w_m$ when reading the sequence $x_{1}^{m}$ with the language model.
Next, we calculate the difference $\Delta_n$ between the activation values in the normal and repetition ranges as a score to quantify the effect of neurons on repetition:
\begin{align}
\Delta_n = \bar{a}_n - a_n. 
\end{align}
Here, larger $\Delta_n$ means the neuron $n$ are activated more strongly in the repetition range than the normal range.
We define the top $K$ neurons with the largest $\Delta_n$ as repetition neurons for the model $\theta$.

\subsection{Observation of Repetition Neuron}
\label{sec:observation}
Figure \ref{fig:delta} shows the obtained $\Delta_n$ of all neurons, sorted in ascending order, for four language models.
It is evident that only a small number of neurons exhibit remarkably high $\Delta_n$ values.
This distribution is consistent with existing reports, which suggest that neuron activation is typically sparse~\cite{li2023the,voita2023neurons}.
This also indicates that only a limited number of repetition neurons are activated exclusively in the repetition range.

Figure \ref{fig:averagedActivation} shows the average activation values of the top four repetition neurons in Gemma-2B, measured across 1,000 texts for 30 tokens before and after the beginning of repetition.
As repetition continues, the activation values of these neurons increase.
This finding suggests that the repetition neurons respond to the recurrence of input tokens.
We hypothesize that when the repetition neurons are strongly activated, the model starts to interpret copying previous tokens as a task, thereby falling into repetition (see \S \ref{sec:intervention}).

Figure \ref{fig:neuronInLayers} presents the distribution of repetition neurons across different layers.
The last layer contains the largest number of repetition neurons in all models, while a secondary peak appears in the intermediate layers.
This suggests the existence of two types of repetition neurons: those that detect repeating patterns in the intermediate layers and those that drive the model to replicate previous contexts in the uppermost layer.
The presence of repetition neurons in both the final and intermediate layers aligns with previous findings that task-specific neurons tend to reside in higher layers~\cite{wang2022finding}, and task-related parameters and hidden states are often found in intermediate layers~\cite{hendel2023context,merullo2024circuit}.
Figure \ref{fig:layerDiffX} and \ref{fig:layerDiffR} in Appendix \ref{sec:ablation} show that the location patterns of the repetition neurons remain consistent across variations in hyperparameters $|X|$ and $r$.

\section{Intervention to Repetition Neurons}
If our hypothesis that the repetition neurons invoke repetition is correct, we should be able to control the repetition problem by intervening with these neurons~\cite{arora2018stronger,wang2022finding}.

\label{sec:intervention}
\begin{figure}[t]
\centering
\includegraphics[width=7.5cm]{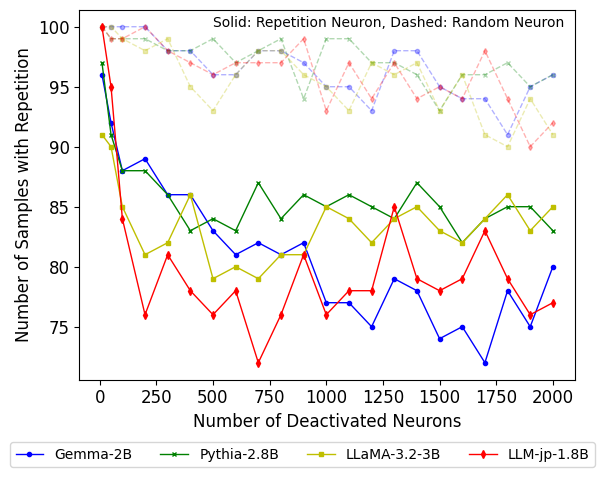}
\caption{
    The number of samples with repetition after deactivating the repetition neurons for the texts originally with repetition.
}
\label{fig:prevention}
\end{figure}

\begin{figure}[t]
\centering
\includegraphics[width=7.5cm]{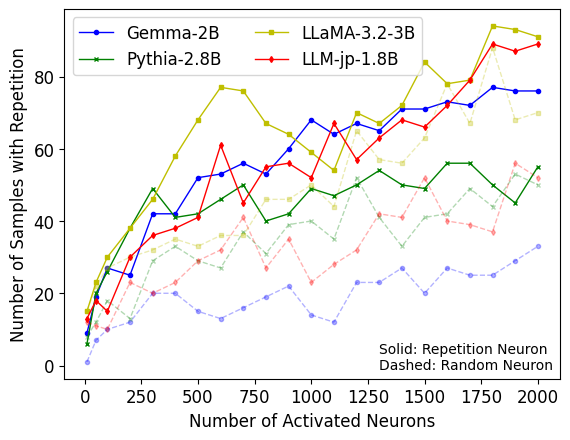}
\caption{
    The number of samples with repetition after activating the repetition neurons for the texts originally without repetition.
}
\label{fig:invoking}
\end{figure}

\begin{table*}[t]
\centering
\footnotesize
\begin{tabular}{p{7.5cm}|p{7.5cm}}
\hline
\textbf{Original Greedy Output}  & \textbf{Intervened Greedy Output} \\ \hline
\transparent{1.0}\ctext[RGB]{255,255,255}{The latest trend in design for the kitchen sink drain is the use of a stainless steel sink drain. This is a great way to add a touch of class to your kitchen. Stainless steel sinks are also very durable and easy to clean.  The $\triangleright$ stainless steel sink drain is a great way to add a touch of class to your kitchen. It is also very durable and easy to clean. What is a stainless steel sink drain? A stainless steel sink drain is a type of sink drain that is made from stainless steel. Stainless steel is a type of metal that is resistant to corrosion and rust.} & \transparent{1.0}\ctext[RGB]{255,255,255}{The latest trend in design for the kitchen sink drain is the use of a stainless steel sink drain. This is a great way to add a touch of class to your kitchen.} \ctext[RGB]{251,227,214}{Stainless steel sinks are also very durable and easy to clean.  The $\blacktriangleright$ stainless steel sink drain is a great way to add a touch of class to your kitchen.} \ctext[RGB]{219,242,209}{Stainless steel sinks are also very durable and easy to clean. The stainless steel sink drain is a great way to add a touch of class to your kitchen.} \ctext{202,238,251}{Stainless steel sinks are also very durable and easy to clean. The stainless steel sink drain ...} \\\hline
\end{tabular}
\caption{
    The example of generation by Gemma-2B with and without intervention to the repetition neurons. 
    $\blacktriangleright$ indicates the beginning point of the intervention to invoke the repetition.
    We also indicate this point in the original greedy output with $\triangleright$ for visibility.
    Color-boxes show the repeating phrases.
}
\label{tab:invokingExample}
\end{table*}

\subsection{Preventing Repetition}
\label{sec:prevention}
\noindent\textbf{Setup:}
In this section, we test whether deactivating the repetition neurons can more effectively suppress repetition compared to deactivating randomly selected neurons.
For this experiment, we generated an additional unseen 100 texts containing repetition for each language model, using the same method described in \S \ref{sec:dataset}.
We then deactivated the repetition neurons by setting their activation values to $0.0$, starting from the token where the original text begins to repeat (e.g., in the case of Figure \ref{fig:averagedActivation}, from the generation step of second ``H'').

\noindent\textbf{Result:}
Figure \ref{fig:prevention} shows the number of samples containing repetition after deactivating varying numbers of repetition neurons (solid lines) compared to randomly selected neurons (dashed lines).
As the figure demonstrates, deactivating the repetition neurons effectively reduces the number of samples with repetition compared to deactivating randomly selected neurons.
This result confirms that the repetition neurons identified by our method are indeed responsible for causing the repetition problem.
We observed that deactivating repetition neurons reduces the number of samples with repetition by up to 25\% (and by as much as 35\% with optimal hyperparameter settings, as shown in Figures \ref{fig:deactivationDiffX} and \ref{fig:deactivationDiffR}).
This suggests that roughly 30\% of the repetition problem can be attributed to the repetition neurons.
Table \ref{tab:reducedExamples} in \S \ref{sec:additional_case_study} provides an example where repetition was successfully suppressed, illustrating that the generated text remains grammatically coherent despite neuron intervention. 
Besides, the perplexity is not largely damaged by deactivating the repetition neurons, as shown in Figure \ref{fig:pplDeactivate}, which supports the coherency of the performance quantitatively.
This confirms that the repetition neurons are specifically responsible for triggering repetition.

\subsection{Invoking Repetition}
\label{sec:invoking}
\noindent\textbf{Setup:}
In contrast to the experiment in \S \ref{sec:prevention}, this section investigates whether activating the repetition neurons leads the model to produce repetitive outputs more effectively than activating randomly selected neurons.
We newly prepared 100 unseen samples for each language model that do not contain repetition.
Each sample consists of 210 tokens, with the first 10 tokens generated randomly and the remaining tokens generated greedily.
Similar to the experiments in  \S \ref{sec:prevention}, we forcibly activate the repetition neurons starting from the 51st token during the generation process.
The neurons are activated by adding $1.0$ to their original activation values.

\noindent\textbf{Result:}
Figure \ref{fig:invoking} presents the number of samples exhibiting repetition after activating repetition neurons and randomly selected neurons.
The figure demonstrates that repetitive samples increase as more neurons are activated.
Furthermore, the activation of repetition neurons is more effective at invoking repetition compared to the activation of randomly selected neurons.
Figure \ref{fig:pplActivate} also demonstrates that activating repetition neurons significantly worsens perplexity, suggesting an increased likelihood of generating repetitive tokens.
These results support our hypothesis that neurons with higher $\Delta_n$ function as ``skill neurons'' that trigger repetitive behavior.
Activating randomly selected neurons also leads to many repetitive samples, suggesting that factors like unstable hidden states also contribute to the repetition problem in addition to the repetition neurons.

\noindent\textbf{Case Study:}
Table \ref{tab:invokingExample} provides a typical generation example obtained by activating the repetition neurons.
The table highlights the text range where we forcibly activate the repetition neurons with the bold font.
Interestingly, the model does not immediately begin repeating tokens following the intervention.
Instead, once it completes the sentence it is generating, the model starts to replicate text that appeared \textbf{before} the point of intervention.
This suggests that the repetition neurons encourage the model to copy previous outputs rather than simply generating tokens that are easily repeated.

\section{Comparison with Induction Heads}
\label{sec:heads}

\begin{figure*}[t]
\centering
\includegraphics[width=\textwidth]{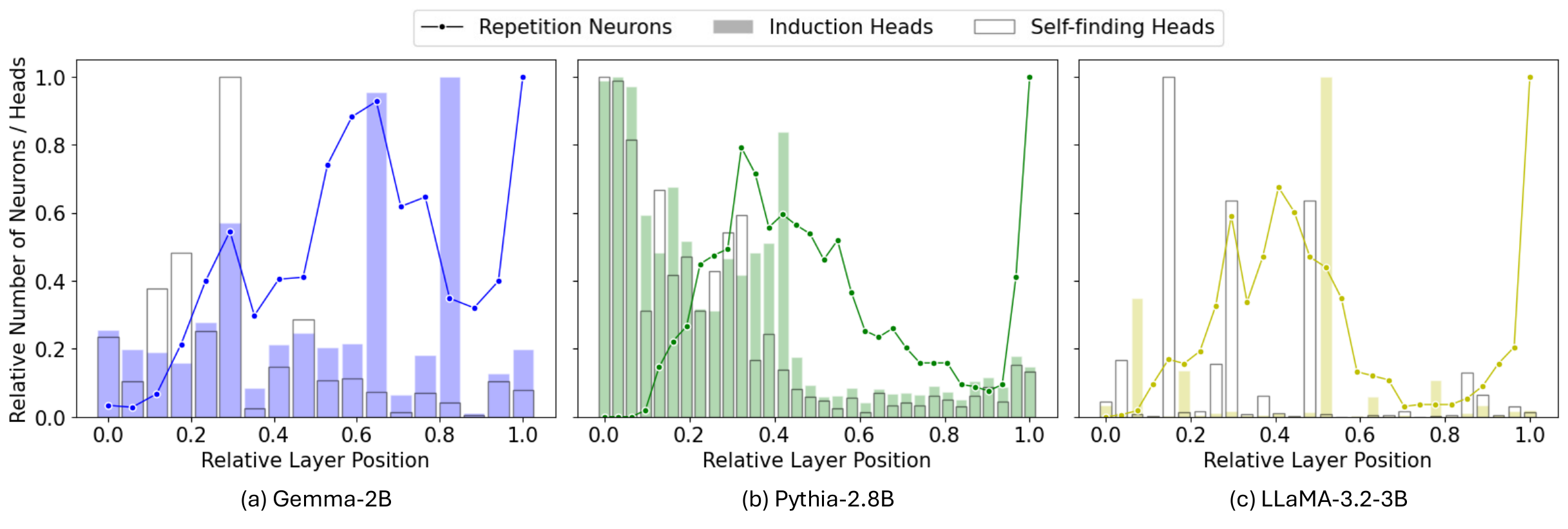}
\caption{
        Frequency of reputation neurons (lines), induction heads (colored bars), and self-finding heads (edged bars) for repetition over three English models.
    }
\label{fig:neuron-head}
\end{figure*}

\begin{figure}[t]
\centering
\includegraphics[width=7.5cm]{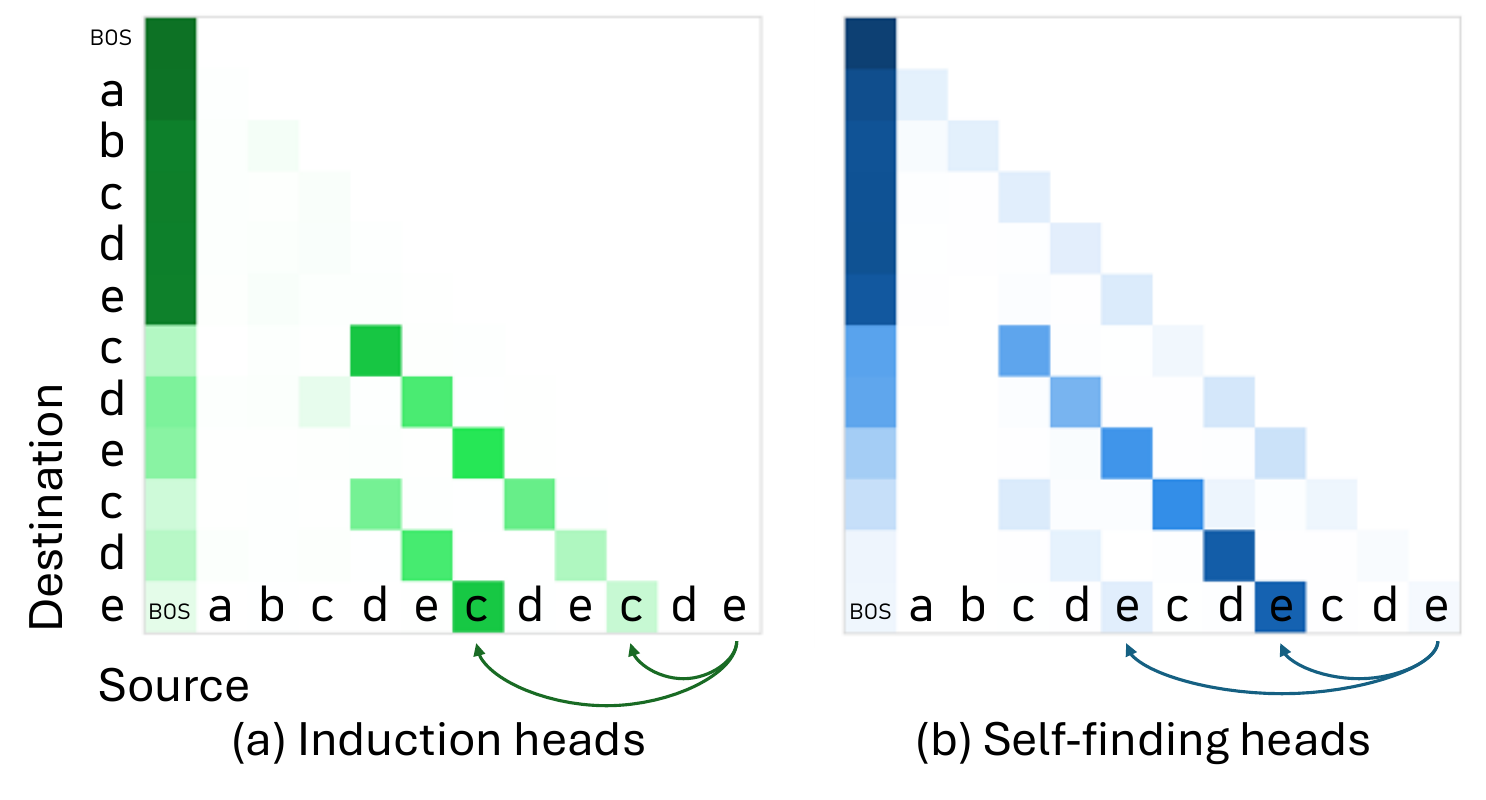}
\caption{
    Examples of attention heat-maps for induction and self-finding heads capturing repetition inputs: ``a b c d e c d e c d e''.
    The expected next token is ``c''.
}
\label{fig:attn_example}
\end{figure}

Several works in in-context learning have examined how attention heads, particularly induction heads ~\cite{olsson2022context,bansal2023rethinking,crosbie2024induction}, exhibit repetitive behaviors. 
These studies explore how in-context learning performance relates to an LLM’s ability to copy patterns in synthetically generated repeating sequences. 
Building on these insights, we focus on a neuron-based analysis of repetition actually generated by the LLMs themselves. 
In this section, we compare the “repetition neurons,” “induction heads,” and “self-finding heads” derived from the same repeating texts using three LLMs: Gemma-2B, Pythia-2.8B, and LLaMA-3.2-3B.

Figure \ref{fig:neuron-head} shows the distribution of the repetition neurons (the same results as in Figure \ref{fig:neuronInLayers}), induction heads, and self-finding heads over layers. 
In our analysis, we define “induction heads” as heads attending to repeating tokens that are to be generated after the current input token (Figure \ref{fig:attn_example}a). 
We also define “self-finding heads” as heads attending to the repeating token identical to the input token (Figure \ref{fig:attn_example}b). 
We identified a head as induction or self-finding if its total attention score for the target tokens that appear after the second repeating position exceeds 0.5. 
We then summarize their layer positions to see how these heads align with repetition neurons.

The observed behavior varies by model architecture. As shown in Figure \ref{fig:neuron-head}a, Gemma-2B’s repetition neurons share two peaks with the induction heads, and one of these peaks is also shared by self-finding heads. This suggests that certain repetition neurons are activated in response to both induction and self-finding heads capturing repetition. However, the highest induction-head peak (layer 14 of 18) does not coincide with the highest repetition-neuron peak (layer 18 of 18).

Figures \ref{fig:neuron-head}b and \ref{fig:neuron-head}c present a different pattern for Pythia and LLaMA, where we do not observe a strong alignment between repetition neurons and induction heads. Nevertheless, similar to Gemma-2B, some peaks in the early layers of repetition neurons correspond to peaks of self-finding heads. This suggests that repetition neurons respond to self-finding patterns in earlier layers and take on different roles in later layers.

Overall, this comparison among repetition neurons, induction heads, and self-finding heads reveals coordinated interactions while showing their distinct roles in detecting and invoking repetition.

\section{Conclusion}
We proposed a method to identify the repetition neurons that contribute to the repetition problem in text generation.
These neurons are located in both the intermediate and final layers of the Transformer, similar to skill neurons and task vectors.
Our experimental results show that by intervening in the activity of these repetition neurons, we can control the occurrence of repetitive outputs.

\section*{Limitations}

The primary goal of this short paper is to report the existence of repetition neurons in repetitive texts and to describe their basic behavior.
We recognize that our findings are likely to spark further discussion, which lies beyond the scope of this work.
To facilitate future research, we outline several key topics related to repetition neurons:
\begin{itemize}
\item We prepared the dataset without considering detailed aspects of repetition (\S \ref{sec:dataset}), such as the length of each repetitive phrase. By focusing on specific phrase lengths, can we identify particular tendencies in the behavior of repetition neurons?
\item We observed two distinct peaks in the distribution of repetition neurons across layers in Figure \ref{fig:neuronInLayers}. What are the functional differences between neurons located in the intermediate layers and those in the final layer?
\item The experimental results of deactivating the repetition neuron suggest that roughly 30\% of the repetition problems are caused by the repetition neuron (\S \ref{sec:prevention}). What causes the rest 70\% repetition problem?
\item Does the behavior of repetition neurons change against the model configuration (e.g., the parameter size, the language used in the pre-training, the activation functions, and so on)?
\item We used the simple intervention to the repetition neurons: replacing the activation value with $0.0$ for deactivation (\S \ref{sec:prevention}) and adding $1.0$ for activation (\S \ref{sec:invoking}). What can we observe when gradually increasing or decreasing the activation value instead of the simple replacement or addition?
\item Beyond the neuron-based and head-based analysis (\S \ref{sec:heads}), can we find any other specific circuit in the LLMs' calculation when outputting the repetitive texts?
\end{itemize}
Some of the above topics are partially discussed in the appendix.
We believe that our findings in this paper help the further discussion to reveal the inner working of the repetition problem.

\section*{Acknowledgement}
This work was supported by JST, CREST Grant Number JPMJCR20D2, Japan.

\bibliography{custom}

\clearpage
\appendix

\section{Comparison with Existing Work}
\label{sec:comparison}
The concept of the ``repetition neuron'' was first introduced by \newcite{wang2024mitigating}, where they showed its impact on machine translation using in-context learning. 
While our research was conducted concurrently and independently, the fact that multiple research teams are exploring similar topics underscores the growing interest in understanding repetition within NLP models. 
Our findings in general text generation complement their results in the machine translation task, supporting the idea that repetition neurons play a broader role across various generation tasks. 
Below, we outline the key distinctions between our work and theirs to highlight our unique contributions.

Unlike their focus on improving performance in in-context learning for machine translation by editing repetition neurons, our research aims to uncover the inner workings of LLMs when processing repetitive text, specifically from the perspective of repetition neurons. 
To achieve this, we employed four different pre-trained language models (three English, one Japanese) and demonstrated that repetition neurons are not restricted to a single architecture on a specific task like machine translation with LLaMA-7B but are observable across various architectures (\S \ref{sec:models}, \ref{sec:dataset}). 
Our broader focus on general text generation highlights the versatility of the repetition neuron phenomenon, as compared to the task-specific nature of the machine translation context used in \newcite{wang2024mitigating}.

Our experiments also provide a more detailed analysis of the distribution of repetition neurons across layers in different models and under varying hyperparameters, something not covered in previous work (\S \ref{sec:observation} and \S \ref{sec:ablation}). 
While both studies involve deactivating repetition neurons to observe the impact on generation, our experiments present a comprehensive comparison across four language models, revealing performance changes as the number of deactivated neurons varies (\S \ref{sec:prevention}). 
One insight that emerges from our findings is that selecting only the top 300 neurons, as in \newcite{wang2024mitigating}, may be insufficient for models of larger scale, a point we explore in depth. 
In addition, our exploration of neuron activation to deliberately induce repetition (\S \ref{sec:invoking}) introduces a novel dimension to this research.

Methodologically, our approach to identifying repetition neurons by comparing activation values before and after the repetition point is more straightforward than their attribution score-based method~\cite{dai2022knowledge}. 
Given that both methods yield similar outcomes in terms of controlling repetition, our simpler approach could serve as an alternative for identifying repetition neurons in large-scale models.

In sum, while our findings do not conflict with those of \newcite{wang2024mitigating}, our work complements their research by providing a broader, more detailed exploration of the role of repetition neurons. 
Our findings not only validate the existence of these neurons across different architectures but also contribute novel insights into their layer-wise distribution and activation patterns. 
These insights pave the way for more targeted interventions in controlling repetition across various language generation tasks.

\section{Example of Generated Repetition}
\label{sec:example_of_dataset}
Table \ref{tab:examples} shows the actual examples of the dataset created in the manner explained in \S \ref{sec:dataset}.
In this table, the bold font highlights the repeated phrases.
Note that here we highlight the text range from the first repeating phrases to the end of the third repeating phrase, while the repetition range mentioned in \S \ref{sec:method} refers to the span after the second repeating point.
As shown in this table, there are various lengths of repeated phrases in each sample.
Future work should focus on the effect of differences in the repetition style on the repetition neurons.

\begin{table*}[]
\centering
\footnotesize
\begin{tabular}{p{15cm}}
\hline\hline
\textit{\textbf{Gemma-2B}} \\ \hline
... ask this, but I’m curious about the black hat. I’ve been looking for a good one for a while now, but I’ve never found one that I really liked. I’ve been looking at the black hat, but \textbf{I'm not sure if it's the right one for me. I'm not sure if it's the right one for me. I'm not sure if it's the right one for me. ...} \\ \hdashline
... come to me.” Smith said she has seen a rise in the number of students who are dealing with mental health issues. “I think it’s because of the pandemic,” she said. “I think it’s because of the isolation\textbf{. I think it’s because of the stress of the world. I think it’s because of the stress of the world. I think it’s because of the stress of the world ...} \\ \hdashline
... I have been using the same starter for the last 2 months. I have been using the same flour and water. I have been using \\
the same method. I have been using the same oven. I \textbf{have been using the same technique. I have been using the same everything. I have been using the same everything. I have been using the same everything. I ...}  \\ \hline
\textit{\textbf{Pythia-2.8B-Deduped}} \\ \hline
... good. I did add a little more salt than the recipe called for. I also added a little more pepper. I also added a little more 
garlic powder. I also added a little more onion powder. I also added a little more oregano\textbf{. I also added a little more basil. I also added a little more basil. I also added a little more basil ...} \\ \hdashline
... a bad mood. "I'm sorry," I said. "I didn't mean to upset you." "It's okay," she said. "I'm just glad you're here." "I'm glad 
I'm \textbf{here, too." "I'm glad you're here, too." "I'm glad you're here, too." "I'm glad you're ...} \\ \hdashline
... look at the flower and see if it is a little bit different. If it is, you should not be too concerned. If it is a little bit different, 
you should be able to use it. If you are using a new flower\textbf{, you should be able to use it. If you are not, you should be able to use it. If you are not, you should be able to use it. If you are not ...} \\ \hline
\textit{\textbf{LLaMA-3.2-3B}} \\ \hline
... It is a book of the plan of salvation. It is a book of the plan of happiness. It is a book of the plan of mercy. It is a book of 
the plan of redemption. It is a book of the plan of life\textbf{. It is a book of the plan of happiness. It is a book of the plan of happiness. It is a book of the plan of happiness ...} \\ \hdashline
... hesion and proliferation of human breast cancer cells. The CD44 expression was analyzed by flow cytometry and immunohistochemistry. The CD44 expression was also analyzed by reverse transcription-polymerase chain reaction (RT-PCR) and Western blotting\textbf{. The CD44 expression was also analyzed by immunohistochemistry. The CD44 expression was also analyzed by immunohistochemistry. The CD44 expression was also analyzed by immunohistochemistry ... }\\ \hdashline
... ? A) 10 B) 12 C) 15 D) 20 E) 25 Answer: Let x be the number of buses on the route. 21x/4 = 21x/4 * 3/4 \textbf{= 3/4 * 21x = 3/4 * 21x = 3/4 * 21x ... }\\ \hline
\textit{\textbf{LLM-jp-3-1.8B}} \\ \hline
\begin{CJK}{UTF8}{ipxm}...と言って豆まきをします。この豆まきは、「鬼」を追い払うという意味があります。「鬼」は「外」に追い出して、「福」は「内」に招き入れる。つまり、\end{CJK}\begin{CJK}{UTF8}{ipxg}\textbf{「鬼」を追い出して、「福」を招き入れる。「鬼」を追い出して、「福」を招き入れる。「鬼」を追い出して、「福」を招き入れる。...}\end{CJK} \\ \hdashline
\begin{CJK}{UTF8}{ipxm}...第5回: 第4回の続き \#\# 第4回: 第3回の続き \#\# 第3回: 第2回の続き \#\# 第2回: 第1回の続き \#\# 第1\end{CJK}\begin{CJK}{UTF8}{ipxg}\textbf{回: 第0回の続き \#\# 第0回: 第0回の続き \#\# 第0回: 第0回の続き \#\# 第0 ...}\end{CJK} \\ \hdashline
\begin{CJK}{UTF8}{ipxm}... 、 必 要 な 措 置 を 講 ず る こ と を 命 ず る こ と が で き る 。（賃貸人の責めに帰すべき事由による場合の契約解除） 第十三条 賃貸人は、\end{CJK}\begin{CJK}{UTF8}{ipxg}\textbf{賃貸借の期間が満了する前に賃貸借の期間が満了する前に賃貸借の期間が満了する前に ...}\end{CJK} 
\\ \hline\hline
\end{tabular}
\caption{Three exapmles of generated samples with perturbation by each langauge model. This table shows 30 tokens before the begenning of repetition and three times repetition.}
\label{tab:examples}
\end{table*}

\section{Additional Case Study}
\label{sec:additional_case_study}
\begin{table*}[]
\centering
\footnotesize
\begin{tabular}{p{7.5cm}|p{7.5cm}}
\hline
\textbf{Original Greedy Output} & \textbf{Intervened Greedy Output} \\\hline
What's the one thing you love the most about your home? Is it the view? The location? The size? The layout? The design? The style? The architecture? The materials? The colors? The finishes? The fixtures? The appliances? The lighting? The furniture? The art? The decor? The landscaping? The pool? The fireplace? The kitchen? The bathroom? The closet? The storage? The garage\ul{? The driveway? The driveway? The driveway? The driveway? The driveway? The driveway? The driveway? The driveway? The driveway? The driveway? The driveway? The driveway? The driveway? The driveway? The driveway? The driveway? The driveway? ...} & What's the one thing you love the most about your home? Is it the view? The location? The size? The layout? The design? The style? The architecture? The materials? The colors? The finishes? The fixtures? The appliances? The lighting? The furniture? The art? The decor? The landscaping? The pool? The fireplace? The kitchen? The bathroom? The closet? The storage? The garage? The driveway\textbf{? If you're like most of us, you probably have a million and one things you love about your home. But, if you're like me, you also have a million and one things you don't like about your home. I've lived in my 1950s-era home for ...}
\\ \hline
\end{tabular}
\caption{
    An example of text generation that repetition is reduced by deactivating repetition neurons (Gemma-2B).
    The repeated texts are highlighted with underline in the original greedy output.
    In the intervened output, bold indicates the range where the repetition neurons are deactivated.
}
\label{tab:reducedExamples}
\end{table*}
Table \ref{tab:reducedExamples} shows an example of text generation with Gemma-2B, where we deactivate the top 600 repetition neurons.
The original greedy generation falls into the repetition of the short phrase ``The driveway?'' after listing similar phrases.
By deactivating the repetition neurons, the model terminates to list similar phrases and begins to generate natural sentences.
This result implies that some repetition neurons have an effect of making the model copy the template ``the \_\_\_?'' and some other neurons are in charge of copying ``driveway''.

\begin{figure*}[ht]
    \begin{subfigure}[b]{0.48\textwidth}
        \includegraphics[width=\textwidth]{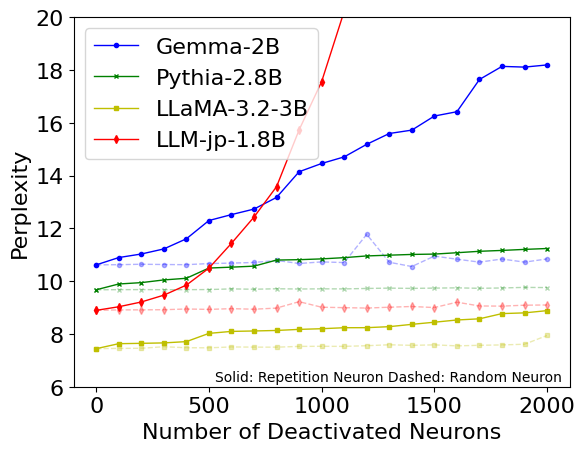}
        \caption{Deactivation}
        \label{fig:pplDeactivate}
    \end{subfigure}
    \hfill
    \begin{subfigure}[b]{0.48\textwidth}
        \includegraphics[width=\textwidth]{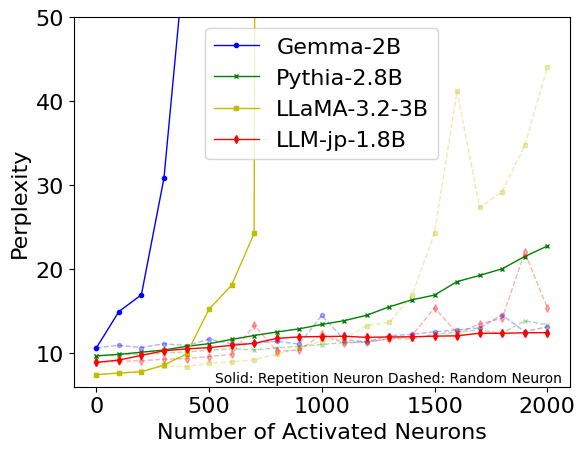}
        \caption{Activation}
        \label{fig:pplActivate}
    \end{subfigure}
    \caption{Perplexity on the test split of WikiText-2 with the intervention to repetition neurons.}
\end{figure*}

\section{Perplexity with Intervention}
Figure \ref{fig:pplDeactivate} and \ref{fig:pplActivate} show the changes in perplexity based on different numbers of intervened repetition neurons. 
We used the test split of WikiText-2 for all models, including LLM-jp-1.8B, which was fine-tuned on Japanese corpora.
As shown in Figure \ref{fig:pplDeactivate}, the performance degradation caused by deactivating repetition neurons is relatively moderate. 
Although the impact on perplexity is smaller when intervening on randomly sampled neurons, considering that GPT-2's perplexity on the same corpus was 18.34~\cite{radford2019language}, the degradation caused by deactivating repetition neurons is acceptable.
This suggests that repetition neurons do not significantly affect the generation of normal texts that do not contain repetition. 

Unlike the English models, LLM-jp-1.8B's perplexity increases substantially even when a smaller number of repetition neurons are deactivated. 
This result implies that repetition neurons may be language-specific. 
In other words, neurons identified as repetition neurons in Japanese may serve a different role in English texts, leading to more significant harm to perplexity on English test sets.

In contrast, the perplexity increases dramatically when repetition neurons are activated. 
For instance, the perplexity of Gemma-2B and LLaMA-3.2-3B exceeds 100 when 500 and 800 repetition neurons are activated, respectively, indicating that the model becomes severely impaired with the activation of a large number of these neurons. 
This suggests that repetition neurons play an important role in generating non-grammatical outputs, and their improper activation increases the likelihood of tokens reappearing from earlier in the text.
On the other hand, LLM-jp-1.8B’s perplexity remains largely unaffected by activating its repetition neurons. 
This further suggests that repetition neurons could be language-specific, as those found in Japanese texts do not have a significant impact on the perplexity of English texts.

\begin{figure*}[ht]
    \begin{subfigure}[b]{0.48\textwidth}
        \includegraphics[width=\textwidth]{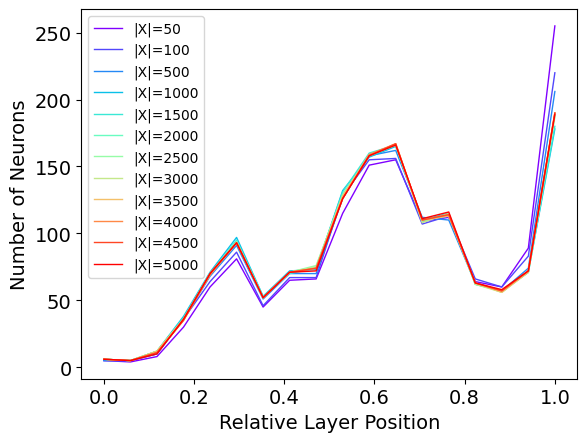}
        \caption{Difference against $|X|$}
        \label{fig:layerDiffX}
    \end{subfigure}
    \hfill
    \begin{subfigure}[b]{0.48\textwidth}
        \includegraphics[width=\textwidth]{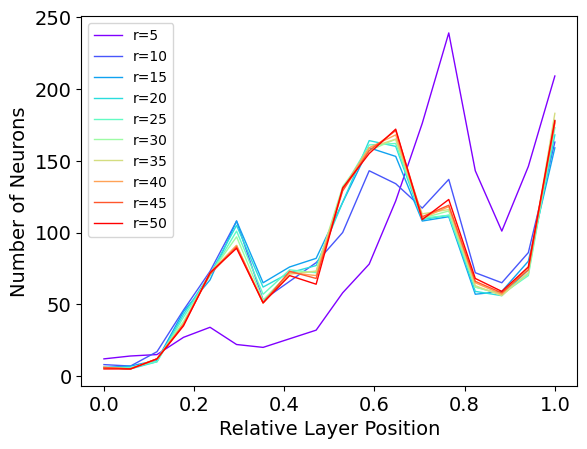}
        \caption{Difference against $r$}
        \label{fig:layerDiffR}
    \end{subfigure}
    \caption{The number of repetition neurons for each layer with various hyperparameters (Gemma-2B).}
\end{figure*}

\section{Ablation Study}
\label{sec:ablation}
The proposed method to seek the repetition neurons has two hyperparameters: the number of repetitive texts generated for the dataset $|X|$ (\S \ref{sec:dataset}) and the text range $r$ to be focused on when calculating activation scores (\S \ref{sec:method}).
In the main body of this paper, we used $|X|=1,000$ and $r=30$.
Herein, we investigate the effect of these hyperparameters on the same experiments using Gemma-2B.
The scope of this ablation study is $|X|=\{50, 100, 500, 1000, 1500, ..., 5000\}$ and $r=\{5, 10, 15, ..., 50\}$.
When investigating the various $|X|$, we fix the other hyperparameter as $r=30$, while $|X|=1,000$ for the investigation of $r$.

Figure \ref{fig:layerDiffX} and \ref{fig:layerDiffR} show the location of the repetition neuron on the Transformer layers (\S \ref{sec:observation}).
As shown in the figure, the size of the dataset $|X|$ does not have large effect on the distribution, which means we can obtain the similar set of repetition neurons both with smaller and larger sizes of datasets.
On the other hand, $r$ has an effect on the distribution to some degree.
For the case of Gemma-2B, we can obtain roughly a similar tendency with $15 \leq r$.

\begin{figure*}[ht]
    \begin{subfigure}[b]{0.48\textwidth}
        \includegraphics[width=\textwidth]{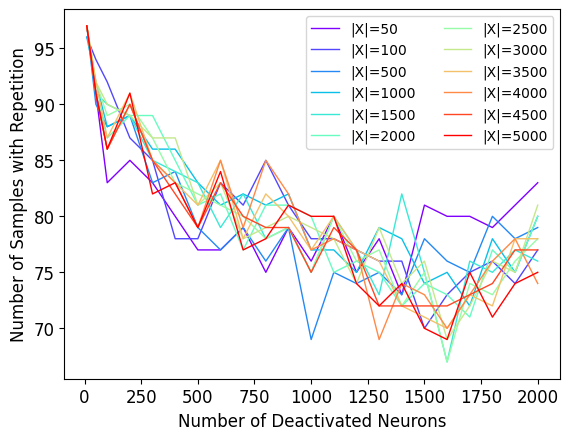}
        \caption{Difference against $|X|$}
        \label{fig:deactivationDiffX}
    \end{subfigure}
    \hfill
    \begin{subfigure}[b]{0.48\textwidth}
        \includegraphics[width=\textwidth]{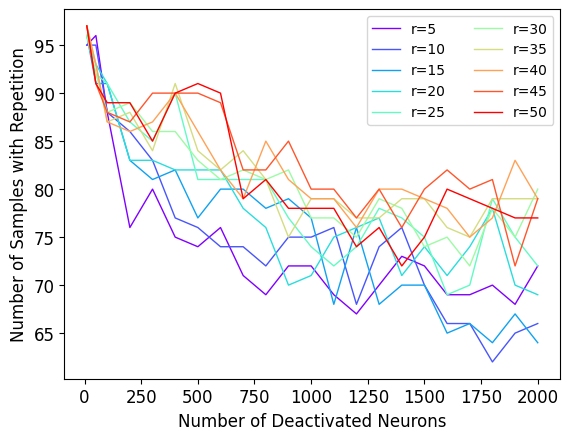}
        \caption{Difference against $r$}
        \label{fig:deactivationDiffR}
    \end{subfigure}
    \caption{The experimental result with deactivating repetition neurons for different hyperparameters (Gemma-2B).}
\end{figure*}

Figure \ref{fig:deactivationDiffX} and \ref{fig:deactivationDiffR} show the difference in the performance for the experiment about deactivating repetition neurons (\S \ref{sec:prevention}).
The experimental result indicates that we can obtain the more reduction effect with the larger number of deactivated repetition neurons.
In contrast, the larger $r$ cause the decrease of the effect to reduce the repetition with larger number of deactivated neurons.
Figure \ref{fig:deactivationDiffR} suggests that $r=10$ or $r=15$ leads to the largest effect on controlling the repetitive generation.

\begin{figure*}[ht]
    \begin{subfigure}[b]{0.48\textwidth}
        \includegraphics[width=\textwidth]{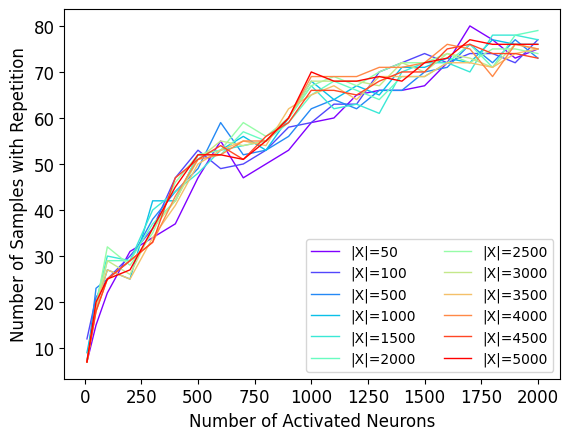}
        \caption{Difference against $|X|$}
        \label{fig:activationDiffX}
    \end{subfigure}
    \hfill
    \begin{subfigure}[b]{0.48\textwidth}
        \includegraphics[width=\textwidth]{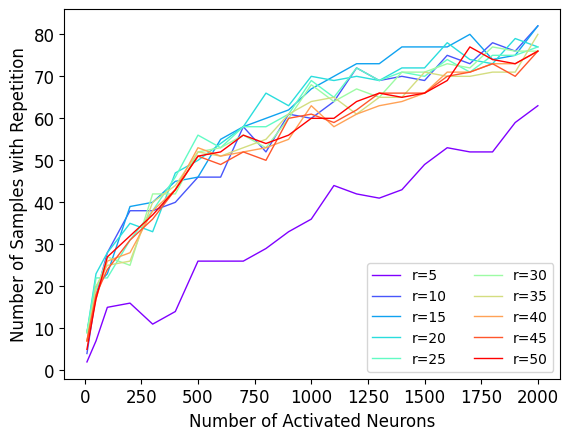}
        \caption{Difference against $r$}
        \label{fig:activationDiffR}
    \end{subfigure}
    \caption{The experimental result with activating repetition neurons for different hyperparameters (Gemma-2B).}
\end{figure*}

The experimental results for the effect of various hyperparameters on the setting with activating repetition neurons in Figure \ref{fig:activationDiffX} and \ref{fig:activationDiffR} show similar trends.
The larger $|X|$ leads to the larger number of repetitive texts while $r=10$ or $r=15$ has the largest effect on the repetitive generation.
These results could be an important clue to investigate the relation between the length of repetitive phrases and the repetition neurons in the future work.

\section{Comparison between Two Model Sizes}
We compared the experimental results with two different sizes of the same architecture: Gemma-2B and Gemma-7B.

\begin{figure}[h]
\centering
\includegraphics[width=7.5cm]{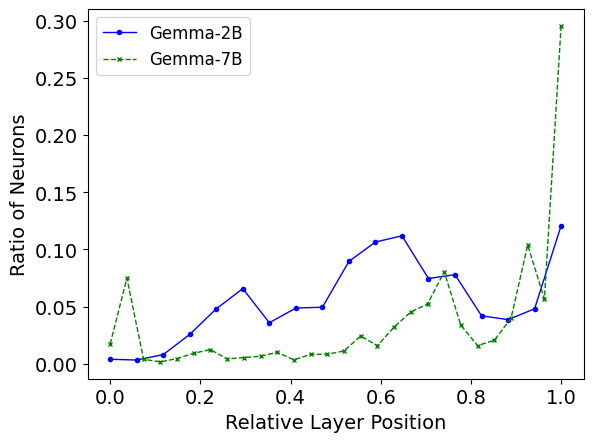}
\caption{
    The number of top 0.5\% repetition neurons for each layer.
    The x-axis shows the relative location of layers against the number of entire layers.
    The y-axis shows the relative number of neurons against the 0.5\% neurons.
}
\label{fig:layer-7b}
\end{figure}

Figure \ref{fig:layer-7b} shows the distribution of repetition neurons over the layers.
Compared to Gemma-2B, the repetition neurons of Gemma-7B are mainly located in the last layer.
This result implies that the nature of repetition neurons varies depending on the size of the language model instead of the architecture.

\begin{figure}[h]
\centering
\includegraphics[width=7.5cm]{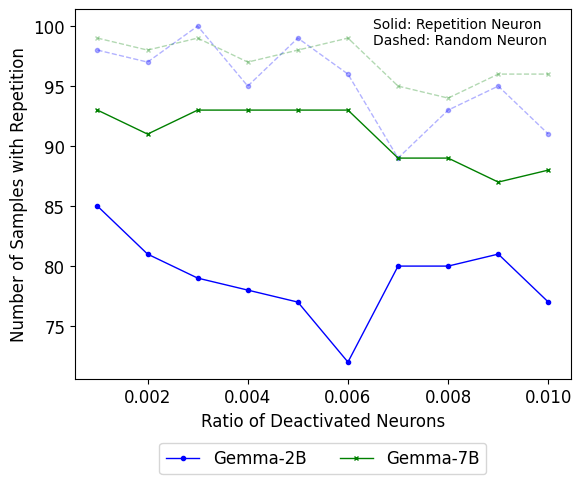}
\caption{
    The experimental results with deactivating repetition neurons for two model sizes.
}
\label{fig:deactivate-7b}
\end{figure}

\begin{figure}[h]
\centering
\includegraphics[width=7.5cm]{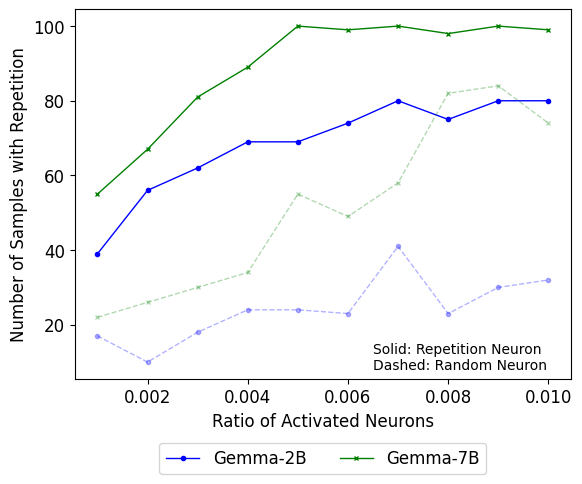}
\caption{
    The experimental results with activating repetition neurons for two model sizes.
}
\label{fig:activate-7b}
\end{figure}

Figure \ref{fig:deactivate-7b} and \ref{fig:activate-7b} show the experimental results of the two experiments with deactivating and activating the repetition neurons, respectively.
The results of Figure \ref{fig:deactivate-7b} indicate that the effect of repetition neurons to prevent the repetition problem becomes smaller when using the larger language model.
This result aligns with the experiment shown in the existing work~\cite{wang2024mitigating}.
In contrast, the activation of repetition neurons of Gemma-7B largely affects the repetitive outputs (Figure \ref{fig:activate-7b}).
These differences in performance show that there is room to be explored about the repetition neuron from broader viewpoints.

\end{document}